\documentclass[sigconf]{aamas} 

\usepackage{balance} %

\setcopyright{none}
    \acmConference[ALA '26]%
    {Proc.\@ of the Adaptive and Learning Agents Workshop (ALA 2026)}%
    {May 25 -- 26, 2026}%
    {Paphos, Cyprus, https://alaworkshop2026.github.io/}%
    {Aydeniz, Delgrange, Mohammedalamen, Yang (eds.)}%
    \copyrightyear{2026}
    \acmYear{2026}
    \acmDOI{}
    \acmPrice{}
    \acmISBN{}
\usepackage[ruled, vlined, linesnumbered]{algorithm2e}
\usepackage{graphicx}
\usepackage{amsfonts}       %

\acmSubmissionID{69}

\title[ReversedQ]{ReversedQ: Opportunities for Faster Q-Learning in Episodic Online Reinforcement Learning}

\author{Sofia R. Miskala-Dinç}
\affiliation{
  \institution{University of Maryland}
  \city{College Park, MD}
  \country{United States}}
\email{smiskala@umd.edu}

\author{Aviva Prins}
\affiliation{
  \institution{University of Maryland}
  \city{College Park, MD}
  \country{United States}}
\email{aviva@umd.edu}

\begin{abstract}

We study model-free $Q$-learning in finite-horizon episodic Markov Decision Processes (MDPs) with stationary dynamics across episodes. We identify a central issue in nascent model-free posterior-sampling works: the reliance on delayed learning in order to prove theoretical guarantees. In particular, we identify three opportunities for faster learning -- (i) value-function update order, (ii) update frequencies, and (iii) value-function initialization. Using \citet{wang2026provably}'s \texttt{RandomizedQ} as a basis, we illustrate these changes and their individual (as well as cumulative) impact in multiple empirical studies. We find that our combined modifications, termed \texttt{ReversedQ}, improve scaled mean cumulative reward compared to \texttt{RandomizedQ}, from 9.53\% to 78.78\% in the Bidirectional Diabolical Combination Lock (BDCL), and from 21.76\% to 61.81\% in a chain MDP. 
\end{abstract}

\keywords{Q-learning, Episodic reinforcement learning, Model-free algorithms, Bayesian exploration} %
         
\newcommand{\BibTeX}{\rm B\kern-.05em{\sc i\kern-.025em b}\kern-.08em\TeX}

\begin{document}

\pagestyle{fancy}
\fancyhead{}

\maketitle 

\section{Introduction}

Reinforcement Learning (RL) algorithms aim to maximize received reward by directing an agent to take actions that affect the environment. These algorithms are useful for solving real-world problems such as inventory control, crop-planning decision support, and robotics \cite{mao2024model, mahajan2024comparative, Kober-2013-7753}. Many such settings are naturally modeled as episodic environments, where interaction proceeds for a fixed number of time steps and then resets. A core challenge in episodic RL is managing the exploration-exploitation tradeoff: to reduce uncertainty about the environment, an agent must sometimes explore actions that may be suboptimal. In order to do this, an algorithm must encourage risk-taking or exploration in some way.

Traditional frameworks predominantly rely on upper confidence bounds, which incentivize exploration via additive bonuses \citep{jaksch2010near, jin2018q, azar2017minimax}. However, these bonuses often prove to be overly conservative. Conversely, posterior-sampling methods--which guide exploration through policy variance derived from a posterior distribution--frequently exhibit superior empirical performance despite a persistent gap in their theoretical characterization \citep{osband2013more, tiapkin2023model}. Recent work by \citet{wang2026provably} made progress in bridging that gap by introducing \texttt{RandomizedQ}, which has both an established sublinear regret bound $\tilde{O}\left(\sqrt{H^5SAK}\right)$ and step-wise policy updates (in contrast to stage-based updates). As a state-of-the-art model-free posterior-sampling algorithm with both a provable regret bound and step-wise updates, \texttt{RandomizedQ} provides a natural baseline for isolating the practical costs of theoretically conservative mechanisms. Demonstrating that its practical performance can be substantially improved is therefore directly relevant to the broader goal of algorithms that are both efficient and deployable.

Although \citet{wang2026provably} correctly prioritize ``agility" (that is, to emphasize responsiveness to new observations), their proposed algorithm overlooks several avenues for enhancing this responsiveness in practical domains. Specifically, we focus on value-function update order, update frequencies, and value-function initialization. We illustrate how these changes may be implemented, using \texttt{RandomizedQ} as a basis, in \texttt{ReversedQ} (Algorithm \ref{alg:reversedq}). Our aim is to foster collaboration on bridging this gap between theory and practice, and to provide intuition for those who work on more complex variants of these systems (i.e., nonstationary or multi-agent systems).

We empirically evaluate the concept of faster learning rates across Bidirectional Diabolical Combination Lock (BDCL) \cite{agarwal2020pc}, and a chain MDP. Our evaluations show that \texttt{ReversedQ} significantly increases total reward, achieving a scaled mean cumulative reward of 78.78\% $\pm$ 0.75 in BDCL and 61.81\% $\pm$ 2.23 in the chain MDP, compared against the baseline's 9.53\% $\pm$ 11.04 and 21.76\% $\pm$ 1.11, respectively. These findings provide strong evidence that our approach offers a marked improvement over \texttt{RandomizedQ}. For practitioners and researchers designing agents for complex, finite-horizon environments, this demonstrates that overcoming the conservative mechanisms required for theoretical regret bounds can yield substantially more agile and efficient exploration in practice.

\section{Related Works}

The central problem in online RL is optimally balancing exploration and exploitation. That is, how to trade off the potential benefit of exploring unknown system dynamics versus leaning into the estimated benefit from past experience. We can divide the approaches in the literature into two groups: bonus-based optimism and Bayesian-based posterior-sampling.

Approximating unknown quantities without sufficient exploration can lead to premature convergence to suboptimal policies. To mitigate this, most provably efficient learning algorithms rely on the \textit{optimism in the face of uncertainty} principle to incentivize exploration \citep{jaksch2010near}. By artificially inflating the perceived value of every state-action pair with an exploration bonus, an upper-confidence bound (UCB) method encourages the learning agent to visit poorly-understood regions of the state-action space. This bonus term decays as uncertainty is resolved. While model-based UCB approaches like \citet{azar2017minimax} and \citet{zhang2025settling} have established minimax-optimal regret bounds (up to log factors), their theoretically derived bonuses are often overly conservative in practice and remain decoupled from the underlying system dynamics.

Rather than relying on additive bonuses, posterior-sampling-based methods maintain a posterior distribution over possible environmental dynamics. Exploration is therefore naturally guided by the variance of policies sampled from this distribution. These Bayesian-based methods have demonstrated strong empirical performance well before their theoretical properties were fully understood \citep{osband2013more, osband2017posterior}. Notably, this gap remains one decade later. \citet{tiapkin2023model}'s empirical evaluations demonstrate that the classic posterior-sampling approach of \citet{osband2013more} can outperform the state-of-the-art UCB-based method in \citet{azar2017minimax}. 

Recent extensions have sought to bridge the theoretical gap by taking a model-free approach: \citet{tiapkin2023model} mimic optimism by mixing $Q$-value estimates to balance exploration and exploitation, and \citet{wang2026provably} extend their work by leveraging ensembles to further calibrate this optimistic bias. Both papers prove a sublinear regret bound of $\tilde{O}\left(\sqrt{H^5SAK}\right)$, where $H$ is the episode length, $S$ is the number of states, $A$ is the number of actions, and $K$ is the number of episodes. In the pursuit of tractable model-free variants, computational compromises have been introduced. While the seminal work in posterior sampling utilized real-time information updates \citep{osband2013more}, recent literature has introduced ``stages" (batched processing of observations) and other techniques that delay the integration of new data. Even nascent frameworks that avoid explicit staging such as \citet{wang2026provably} still exhibit latency in their update cycles. While these developments represent significant theoretical milestones, we will show in this work that ingesting information \textit{quickly} is key to avoiding algorithmic inefficiencies in practice.

\section{Problem Formulation}

We consider a finite, episodic Markov Decision Process (MDP) with horizon $H \in \mathbb{N}$, state space $\mathcal{S}$, and action space $\mathcal{A}$. Episodes are indexed by $k \in [K] := \{1,\dots,K\}$ and within-episode time by $h \in [H]$.  In contrast to standard infinite-horizon formulations where transition dynamics and value functions are time-independent, our finite-horizon setting relies on step-dependent components. Consequently, we denote step-specific variables with a subscript $h$ -- such as policies $\pi_h$, rewards $r_h$, transition probabilities $P_h$, and value functions $V_{h}$ and $Q_h$. At step $h$ of episode $k$, the agent observes $s_{h, k} \in \mathcal{S}$, picks $a_{h, k} \in \mathcal{A}$, receives reward $r_{h, k}(s_{h, k},a_{h, k}) \in [0,1]$, and transitions to $s_{h+1, k} \sim P_{h, k}(\cdot \mid s_{h, k},a_{h, k})$. The episode starts from $s_{1, k}$ and terminates at step $H$. While the reward and transition functions may differ at varying positions within an episode, they are stationary across episodes, i.e., $r_{h,k_1}(\cdot)=r_{h,k_2}(\cdot)$, and similarly for $P_{h,k}(\cdot)$.

A policy is a sequence of decisions $\pi=\{\pi_h\}_{h=1}^H$ where $\pi_h(\cdot \mid s) \in \Delta\mathcal{A}$. For any policy $\pi$, we define the value and action-value functions
\begin{align}
V_{h}^\pi(s) &= \mathbb{E}\!\left[\;\sum_{t=h}^{H} r_{t}(s_{t},a_{t}) \,\middle|\, s_{h}=s \right], \\
Q_{h}^\pi(s,a) &= r_{h}(s,a) + \mathbb{E}\left[\;\sum_{t=h + 1}^{H} r_{t}(s_{t},a_{t}) \,\middle|\, s_{h}=s, a_h = a \right],
\end{align}
respectively, with $V_{H+1}^\pi \equiv 0$. We denote the optimal value functions by $V_{h}^*$ and $Q_{h}^*$, and the optimal policy across all episodes by $\pi^* \in \arg\max_\pi V_{1}^\pi(s_{1})$.

The learner's goal is to maximize the cumulative reward over all episodes, or equivalently, minimize the \textit{regret} under a policy $\pi$ over all episodes:
\begin{equation}
    \text{Regret}_\pi := \sum_{k=1}^{K} \left(V_{1}^*(s_1) - V_{1}^{\pi^k}(s_1)\right).
\end{equation}

\subsection{Bayesian-Based Learning Approach}
Before proceeding, let us introduce the Bayesian-based model-free learning approach to be used throughout. The learner is \emph{model-free}: it observes $(s_{h,k},a_{h,k},r_{h,k},s_{h+1,k})$ during interaction, but does not estimate $P_{h,k}$ or $r_{h,k}$. Rather, it directly estimates $V_{h,k}^\pi$ so as to minimize regret. To do so, the learner samples an ensemble of $Q$-values according to the posterior distribution of the Bayesian model of the unknown MDP system. \texttt{RandomizedQ} is the state-of-the-art algorithm for this problem \cite{wang2026provably}. Therefore, for clarity, we illustrate our ideas using their algorithm as a basis. To indicate terms associated with \textit{exploratory} estimations, we use $\breve{\square}$. To indicate terms associated with \textit{exploitative} estimations, we use $\tilde{\square}$.

\begin{algorithm}[t]
    \caption{ReversedQ}\label{alg:reversedq}
    \KwIn{Inflation coefficient $\kappa$, ensemble size $J$, number of prior transitions $n_0$}
    \KwData{Visitation counter $N_h(s,a)$; Policy Q-table $Q_h(s,a)$; Q-ensembles $\tilde Q_h^j$, $\breve Q_h^j$; value tables $\tilde V_h,\breve V_h$; mixing rate $\eta \gets \frac{1}{\sqrt{H} + 1}$; reward table $rewards$; trajectory table $path$}
    {\color{blue} \tcp{Informed value initialization}} 
    \textbf{Initialize:} $V^0=\{(H-h+1)\}_{h=1}^{H+1}$, $\tilde V_h(s), \breve V_h(s)\gets V_h^0$; $Q_h(s,a), \breve Q_h(s,a), \tilde Q_h^j(s,a),\breve Q_h^j(s,a)\gets V_{h+1}^0; N_h(s,a)\gets 0; $ for any $(j, h, s, a) \in [J]\times [H] \times \mathcal{S} \times \mathcal{A}$ \label{line:init}
    
    \For{episode $k\gets1$ \KwTo $K$}{
      \For{step $h\gets1$ \KwTo $H$}{
      
        Take action $a_h\gets \arg\max_aQ_h(s_h, a)$;\label{line:start_shared1}
        
        Observe $s_{h+1}$ and receive reward $r(s_h, a_h)$
        
        $rewards_h \gets r_h(s_h, a_h)$,
        $path_h \gets (s_h, a_h)$; \label{line:end_shared1}
    
        \BlankLine

        \If{$h = H$}{
            $path[H + 1]\gets (s_{h+1}, a_{arbitrary})$

            \For{$i = H, H - 1,\cdots, 1$} {  \label{line:backward}
                {\color{blue} \tcp{Traverse backwards on trajectory}} 
                $s_i, a_i \gets path_i, r \gets rewards_i$; \label{line:start_shared2}
                
                $\tilde w^j \sim {Beta}(\frac{H + 1}{\kappa}, \frac{N_{i}(s_i, a_i) + n_0}{\kappa})$
                
                $\breve w^j \sim {Beta}(\frac{1}{\kappa}, \frac{N_{i}(s_i, a_i) + n_0}{\kappa})$
    
                $\tilde Q^j_{i}(s_i, a_i) \gets (1-\tilde w^j) \tilde Q^j_i(s_i, a_i) + \tilde w^j(r + \tilde V_{i + 1}(s_{i + 1}))$ \label{line:exploit}
                
                $\breve Q^j_{i}(s_i, a_i) \gets (1-\breve w^j) \breve Q^j_i(s_i, a_i) + \breve w^j(r + \breve V_{i + 1}(s_{i + 1}))$ \label{line:end_shared2}
                
                {\color{gray}$a_{temp} \gets \arg\max_{a \in A}Q_i(s_i, a)$}\
                
                {\color{gray}$j_{temp}\gets \arg\max_{j\in[J]} \breve Q^{\,j}_{i}(s_i,a_i)$}\
                
                $\tilde V_i(s_i) \gets \min \{V^0, \max_{j \in [J]}\tilde Q^j_i(s_i, a_{temp})\}$ \label{line:vtilde}

                {\color{blue} \tcp{Accelerated value updates}} 
                $\breve Q_i(s_i, a_i) \gets \breve Q^{j_{temp}}_i( s_i, a_{temp})$\label{line:qexploreensemble}
                
                $\breve V_i(s_i) \gets \max_{a \in A}\breve Q_i(s_i, a) $ \label{line:vexplore}
                
                $Q_i(s_i, a_i) \gets \min\{Q_i(s_i, a_i), \Bigl( \eta \cdot \max_{j \in [J]}\tilde Q^j_i(s_i, a_i) + \eta \cdot \breve Q_i(s_i, a_i) \Bigr) \}$ \label{line:policyq}
      
                $N_i(s_i, a_i) \gets N_i(s_i, a_i) + 1$
    
            }
            Reset $rewards[1{:}H], path[1{:}H+1]$
        }
      }
    }
\end{algorithm}

\section{Opportunities for Faster Learning}

In this section, we introduce our core conceptual idea for information utilization: encouraging faster learning rates. We call the combination of these three ideas \texttt{ReversedQ}. Building upon \texttt{RandomizedQ} \citep{wang2026provably}, \texttt{ReversedQ} (detailed in Algorithm \ref{alg:reversedq}) integrates three synergistic components: backward passes, accelerated value update schedules, and informed value initialization. Together, these mechanisms allow \texttt{ReversedQ} to propagate observed signals more efficiently. This core idea is surprisingly straightforward: in a stationary system such as ours, there is an advantage to utilizing information as soon as it is received.

\subsection{Backward Pass Through An Episode} 

By construction, in standard RL, the value function ($V_h$) and action-value function ($Q_h$) at step $h\in [H]$ are conditioned on the expected cumulative reward of subsequent time steps. Updating these functions in a forward chronological order introduces delay: updates at step $h$ rely on values from step $h+1$ that have not yet been updated within the current time step. This results in the propagation of old estimates derived from previous, likely suboptimal, trajectories.

In episodic settings, this lag can be mitigated by leveraging the full observability of rewards upon episode completion. We replace \texttt{RandomizedQ}'s forward pass with a backward pass mechanism on line \ref{line:backward}. This ensures that value updates utilize the most recent information available from the current trajectory.

\subsection{Value Update and Restart Timing} 
We suggest changes to the method of updating action-value function $\breve Q_i$ and value table $\breve V_i$, on lines \ref{line:qexploreensemble} and \ref{line:vexplore} respectively, and restarting exploration ensemble $\breve Q^j_i$. These changes allow for agile initial convergence while retaining the adaptability needed to correct for inaccuracies caused by unforeseen failures.

Firstly, staging updates is often beneficial for the purpose of theoretical evaluation of an algorithm. However, it is known that they practically withhold valuable, recent information from the learner \cite{tiapkin2023model}. Naturally, we have allowed the agent to access this information as it becomes available by letting $\breve Q_i(s_i, a_i)$ and $\breve V_i(s_i)$ update per time step (lines \ref{line:qexploreensemble} and \ref{line:vexplore}). 

Secondly, in order to induce and maintain optimism, \texttt{RandomizedQ} periodically resets the optimistically-driven exploration ensemble, $\breve Q_i^j$. While periodic resets can improve adaptability to account for the probability of failure, we have found that indiscriminate resetting hinders learning by discarding valid information, forcing the policy to re-converge at the cost of sample efficiency. In order to balance optimism, we have chosen to no longer reset the exploration ensemble, but to increase the rate at which it is optimistically mixed -- from $(1-\eta)$ to $\eta$ on line \ref{line:policyq}. This allows the agent to continue to benefit from increased optimism without suffering periodic losses. 

Finally, we note that the upper-bound clipping via $\min(\cdot)$ on lines \ref{line:vtilde} and \ref{line:policyq} are inherited from \citet{wang2026provably, wang2026randomizedqcode}'s implementation, where they prevent value estimates from exceeding the initialized upper bound. Although absent in \texttt{RandomizedQ}'s pseudocode, they are present in the released code and carried forward unchanged.

\subsection{Value Initialization} 
A common technique in online RL is the optimistic initialization of $Q$ and $V$ values, which are subsequently refined based on observations. Within the \texttt{RandomizedQ} framework, value resets occur periodically; specifically, the exploratory $\breve{Q}$-ensembles are re-initialized ``to mitigate outdated data and ensure optimism" \cite{wang2026provably}. Unfortunately, \citet{wang2026provably} employ an initial value that exceeds the bounds of the environment to ensure their theoretical bound on regret (Theorem 1). Introducing unnecessary slack reduces learning efficiency. Therefore, we suggest an initial value estimation of $H-h + 1$ instead of $2(H-h + 1)$ (line \ref{line:init}). As we will demonstrate in the following section, even this minor change has a big impact.

\section{Empirical Evaluation}

To evaluate the practical impact of our framework, we empirically evaluate \texttt{ReversedQ} on two MDP systems. All experiments were conducted with the following hyperparameters: number of ensembles $J = 10$, number of prior transitions $n_{0} = \frac{1}{S}$, and inflation coefficient $\kappa = 1$. The tests were performed on a machine with an Intel i7-13700K (3.4 GHz) processor and 64 GB of RAM. 

We analyze \texttt{ReversedQ} and the three components individually, each termed \texttt{RandomizedQ-Backward}, \texttt{RandomizedQ-Update}, and \texttt{RandomizedQ-Init}, against \texttt{RandomizedQ}.

In BDCL, the agent can take $\mathcal{A} = 5$ actions with fail probability $p_{fail} = 0.02$ over an episode of length $H = 5$ and total horizon $T = 2,500$. We compute 20 simulation seeds for each policy evaluation. In the chain MDP, we compute 5 simulation seeds of episode length $H = 50$, horizon $T = 60,000$, and $\mathcal{S} = 20$ states.

\paragraph{Metric}

To measure performance, we compare the scaled mean cumulative rewards, where scaling uses the mean cumulative reward of an Oracle policy as the upper bound and a Random policy as the lower bound:
\begin{equation}
   \text{Scaled}(\pi) = 100 \cdot \frac{\mathbb{E}[R(\pi)] - \mathbb{E}[R(Random)]}{\mathbb{E}[R(Oracle)] - \mathbb{E}[R(Random)]}
\end{equation}

We report 95\% confidence intervals over independent runs.

\subsection{Environments}

\textit{Bidirectional Diabolical Combination Lock} (BDCL) \cite{agarwal2020pc} is an episodic MDP designed to be hard to explore. The MDP has four states: \textbf{Start}, \textbf{Sink} (absorbing), \textbf{Lock 1}, and \textbf{Lock 2}. Each episode begins in \textbf{Start}. Let the action set be $\{0,\dots,A-1\}$ and define $\tau=\lfloor (A-1)/2 \rfloor$; actions $0,\dots,\tau$ route the first step to \textbf{Lock 1}, while actions $\tau+1,\dots,A-1$ route to \textbf{Lock 2} (this handles both even and odd A). Once on \textbf{Lock $i$} at any $t\in\{1,\dots,H-1\}$, exactly one randomly selected ``progress'' action advances along that lock; any other action transitions to \textbf{Sink}, which is absorbing. Rewards are zero in \textbf{Start} and on intermediate lock steps. At $t=H$, a terminal reward $R_i$ is obtained if the agent remains on \textbf{Lock $i$}. The second lock yields higher reward than the first lock ($r_{l1}=0.25$, $r_{l2}=1.0$). The \textbf{Sink} yields $r_{sink}=\frac{1}{8H}$ per step, so an agent spending the entire episode there receives at most $\frac{1}{8}<r_{l1}<r_{l2}$. A failure mechanism with probability $p_{fail}\in[0,1)$ overrides the chosen action and sends the agent to \textbf{Sink} regardless of correctness. The identity of each lock's unique progress action is unfixed and changes across time. Collectively, these design choices make the environment exploration-hard: rewards are sparse and delayed, the progress action is unique at each step, and any deviation -- or failure with probability $p_{fail}$ -- irreversibly transitions the agent to the absorbing \textbf{Sink}.

\begin{figure}[t]
    \centering
    \includegraphics[width=1\linewidth]{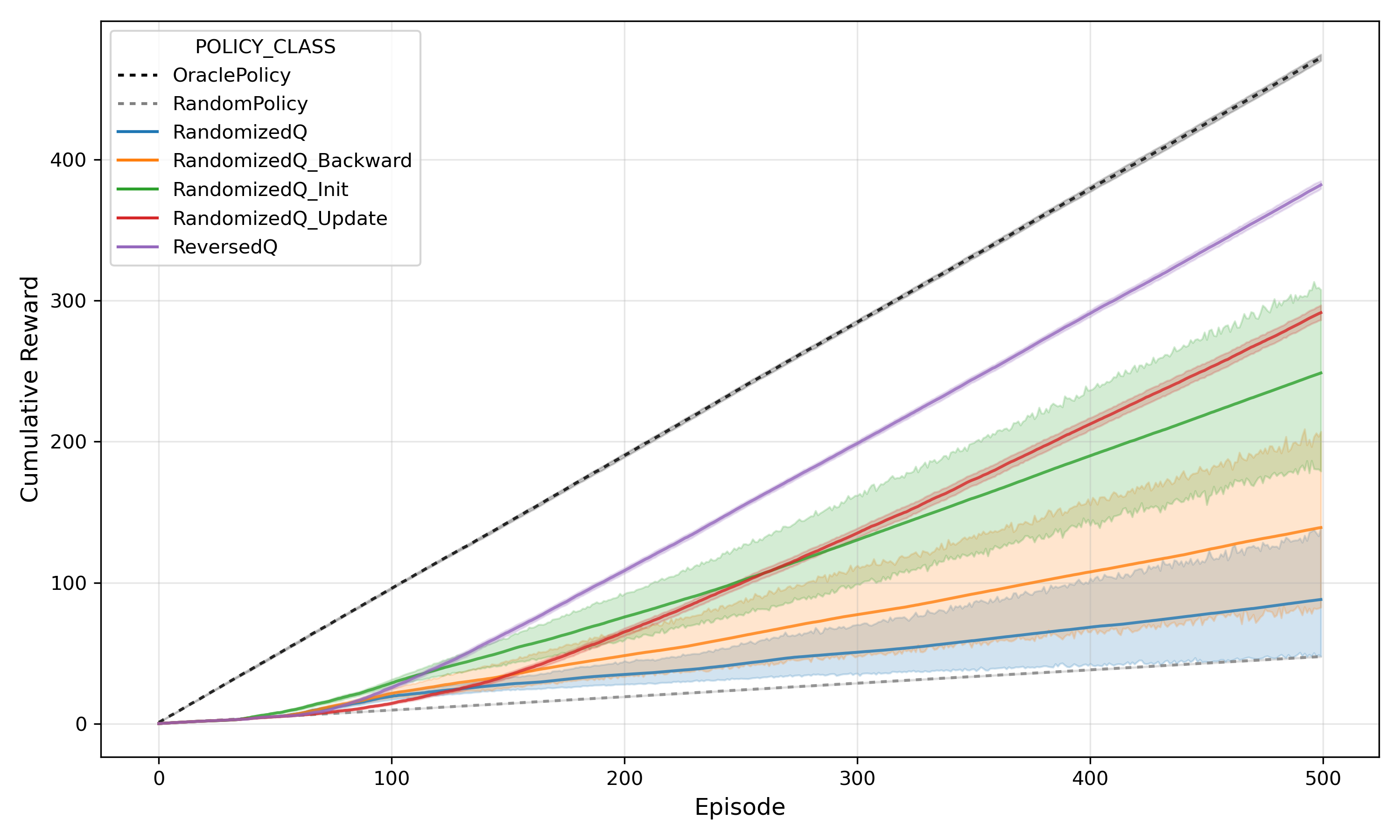}
    \caption{Agents face difficulty exploring BDCL as they encounter several significant, yet suboptimal rewards before exploration to the terminal state in the higher-yielding lock. This creates wide confidence intervals in cumulative rewards across several simulations. However, there is little variance in the reward that \texttt{ReversedQ} achieves, indicating robustness: the policy consistently finds the optimal solution, regardless of slight simulation-to-simulation deviations. Furthermore, an analysis of our individual contributions demonstrates that each component independently benefits algorithmic performance, consistently yielding higher rewards than the baseline, \texttt{RandomizedQ}.}
    \Description{A line plot showing the scaled mean cumulative rewards of five policies across 500 episodes in the BDCL stationary environment. The policies are listed in order of greatest reward to least reward: Oracle, ReversedQ, RandomizedQ_Update, RandomizedQ_Init, RandomizedQ_Backward, RandomizedQ, Random.}
\label{fig:bdcl}
\end{figure}
\begin{table}[t]
\caption{BDCL Scaled Mean Cumulative Reward with 95\% Confidence Intervals}
  \centering
  \begin{tabular}{ll}
    \toprule
    \textit{Policy} & \textit{Scaled Mean Reward} (\%)\\ \midrule
    RandomizedQ & 9.53 $\pm$ 11.04  \\ 
    RandomizedQ-Backward & 21.55 $\pm$ 15.97 \\ 
    RandomizedQ-Init & 47.36 $\pm$ 16.60  \\ 
    RandomizedQ-Update & 57.44 $\pm$ 1.32 \\ 
    \textbf{ReversedQ} & \textbf{78.78} \textbf{$\pm$} \textbf{0.75}\\ \bottomrule
  \end{tabular}
  \label{tab:bdcl}
\end{table}

We also compare against the \textit{chain MDP} environment used in the empirical results of \citet{tiapkin2023model, wang2026provably}. The agent begins in the leftmost state $s_1$. At each step, the agent chooses an action $a \in \{\textbf{left}, \textbf{right}\}$. The chosen action succeeds with probability $p = 0.9$, moving the agent in the intended direction, and fails with probability $1 - p$, moving the agent in the opposite direction. Shifting into a boundary maintains the agent's current position with probability $p$ and moves the agent inward with probability $1-p$. Rewards are sparse: the agent receives rewards only at the boundary states. The starting state $s_1$ yields suboptimal reward $r = 0.05$ and the terminal state $s_S$ grants a large payoff of $r = 1$.

\begin{figure}[t]
    \centering
    \includegraphics[width=1\linewidth]{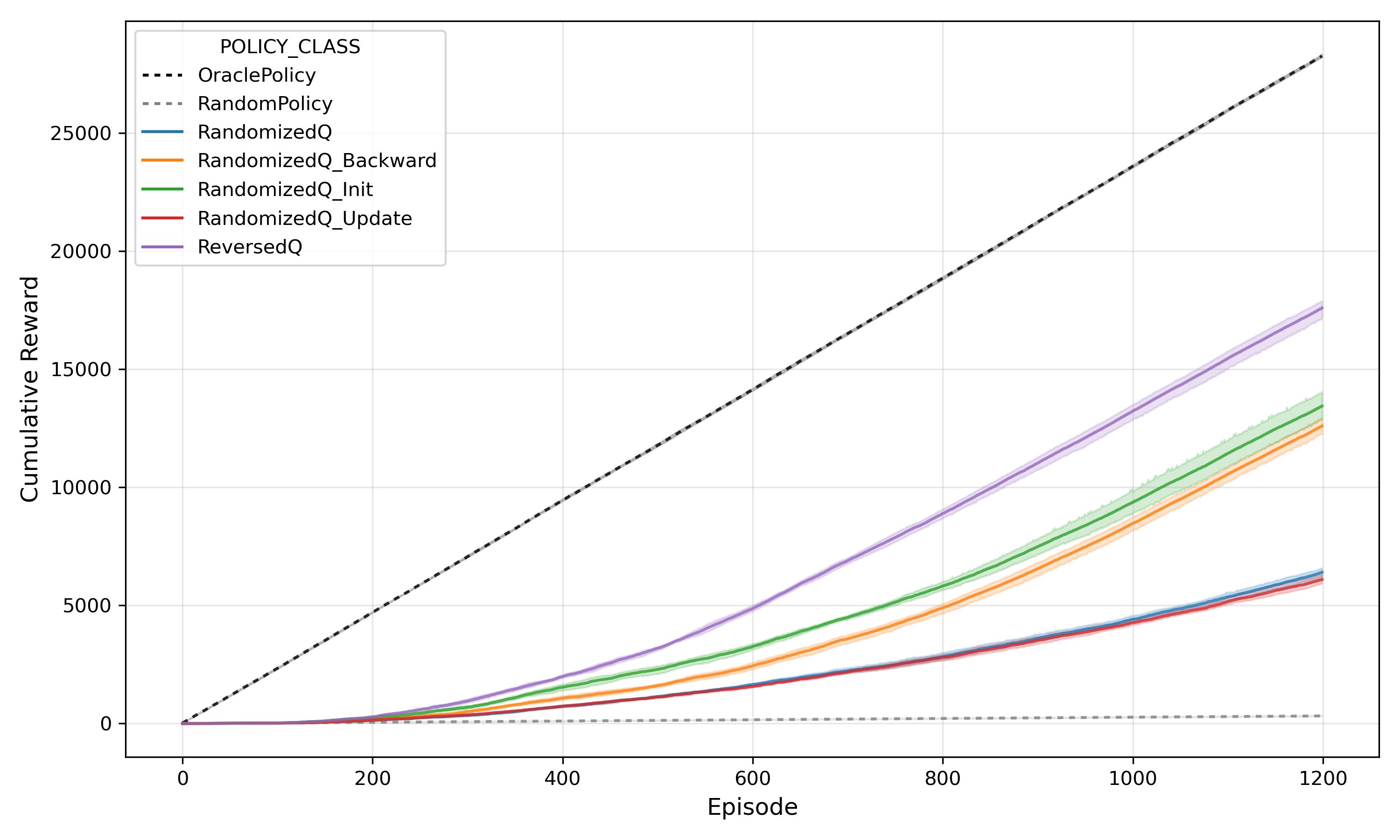}
    \caption{\texttt{ReversedQ} converges in fewer episodes and achieves higher reward than \texttt{RandomizedQ}. Backward passes and informed initialization each improve performance, while the value update and restart timing changes perform similarly to baseline.}
    \Description{A line plot showing the scaled mean cumulative rewards of five policies across 1200 episodes in the chain environment. The policies are listed in order of greatest reward to least reward: Oracle, ReversedQ, RandomizedQ_Init, RandomizedQ_Backward, RandomizedQ_Update,  RandomizedQ, Random.}
    \label{fig:chain}
\end{figure}

\begin{table}[t]
\caption{Chain Scaled Mean Cumulative Reward with 95\% Confidence Intervals}
  \centering
  \begin{tabular}{ll}
    \toprule
    \textit{Policy} & \textit{Scaled Mean Reward} (\%)\\ \midrule
    RandomizedQ & 21.76 $\pm$ 1.11  \\ 
    RandomizedQ-Backward & 43.93 $\pm$ 1.94 \\ 
    RandomizedQ-Init & 46.95 $\pm$ 3.41  \\ 
    RandomizedQ-Update & 20.67 $\pm$ 1.13 \\ 
    \textbf{ReversedQ} & \textbf{61.81} \textbf{$\pm$} \textbf{2.23}\\ \bottomrule
  \end{tabular}
  \label{tab:chain}
\end{table}

\subsection{Results and Discussion}
Our empirical analysis has shown that mitigating theoretically conservative mechanisms enables \texttt{ReversedQ} to achieve significantly higher rewards compared to \texttt{RandomizedQ} across both environments. As seen in Figures \ref{fig:bdcl} and \ref{fig:chain}, \texttt{ReversedQ} takes significantly fewer episodes, and therefore samples, to converge on the optimal policy when compared against \texttt{RandomizedQ} and each isolated component implemented on \texttt{RandomizedQ}.

\paragraph{Backward pass through an episode}
Individual and combined evaluation of this property has shown a significant increase in reward in each scenario. When this change is implemented independently with our baseline algorithm, \texttt{RandomizedQ}, the two together control over-exploration and reduce regret. This leads to hastened convergence and demonstrates that stationary performance can be substantially improved when optimism is calibrated and information from previous steps is fully utilized. In BDCL (Figure \ref{fig:bdcl}) and chain (Figure \ref{fig:chain}) environments, the backward pass property alone improves the scaled reward of \texttt{RandomizedQ} from 9.53\% $\pm$ 11.04 to 21.55\% $\pm$ 15.97 and 21.76\% $\pm$ 1.11 to 43.93\% $\pm$ 1.94, respectively.

\paragraph{Update and restart timing}
Optimized update and restart frameworks allow for better rate and retention of learning. Figure \ref{fig:bdcl} supports this hypothesis, showing improvement in scaled reward from 9.53\% $\pm$ 11.04 to 57.44\% $\pm$ 1.32 in BDCL. However, this change does not significantly impact \texttt{RandomizedQ} individually in the chain environment, yielding roughly baseline performance. The asymmetry between environments is explained by their structure. In BDCL, stale value estimates are particularly costly because a single suboptimal action irreversibly transitions the agent to the absorbing sink state. Prompt updates therefore have a sizable impact. In the chain environment, the reward structure is less complex. Consequently, delayed updates are less detrimental, explaining why this modification alone does not significantly improve over baseline. Despite its limited independent impact in the chain environment, the optimized timing framework is critical; it allows for prompt reactions to recent discoveries and the retention of learning across episodes, which ultimately fosters faster, more stable convergence as seen across the figures.

\paragraph{Value Initialization}
Informed value initialization produces a large and consistent performance gain across both environments. In BDCL, replacing \texttt{RandomizedQ}'s overly optimistic initialization with the tighter bound $V^0_h = H - h + 1$ improves scaled mean reward from 9.53\% $\pm$ 11.04 to 47.36\% $\pm$ 16.60 (Table \ref{tab:bdcl}), indicating that a substantial fraction of simulations accumulate higher rewards much earlier than the baseline. In the chain MDP, the same modification increases performance from 21.76\% $\pm$ 1.11 to 46.95\% $\pm$ 3.41 (Table \ref{tab:chain}). These improvements suggest that \texttt{RandomizedQ}'s original initialization is overly optimistic relative to reachable returns. Correcting excessive optimism can take many episodes, delaying reliable exploration. Initializing at a realistic upper bound keeps beneficial optimism while reducing unnecessary slack, leading to faster calibration. 

\balance
 
\section{Conclusion}

In this work, we show that accelerating the feedback loop between observation and estimation significantly improves posterior-sampling-based online RL. By incorporating backward passes, accelerated value update schedules, and informed value initialization into \texttt{RandomizedQ}, we achieve substantial performance gains across both environments. Specifically, our combined approach, \texttt{ReversedQ}, increased the scaled mean cumulative reward from 9.53\% to 78.78\% in the BDCL environment and from 21.76\% to 61.81\% in the chain MDP environment. Furthermore, our ablation studies confirm that each isolated mechanism independently accelerates convergence and reduces regret.

The empirical performance of \texttt{ReversedQ} substantially exceeds that of \texttt{RandomizedQ} \cite{wang2026provably} across both tested environments. The established regret bound of $\tilde{O}(\sqrt{H^5SAK})$ for \texttt{RandomizedQ} \cite{wang2026provably} represents a worst-case theoretical ceiling; the gap between such bounds and empirical performance aligns with existing observations that posterior-sampling algorithms frequently outperform their theoretical guarantees \cite{osband2017posterior}. Rather than merely reporting this gap, we propose specific algorithmic modifications grounded in a structural explanation of their performance. Moving forward, we would like to formally prove that \texttt{RandomizedQ}’s regret bounds are preserved under these modifications. Furthermore, this framework facilitates the isolation of stationary online RL dynamics from external factors such as non-stationarity or multi-agent interactions. Recent work by \citet{nonaka2025efficient} supports this view: the authors discuss modifying a stationary algorithm for use in a non-stationary setting (indeed, they use \texttt{RandomizedQ} for illustration).

\section*{Acknowledgments}
We gratefully acknowledge Hiroshi Nonaka, Simon Ambrozak, and Amedeo Ercole for their invaluable insights and support. We also thank Dr. William Gasarch and the REU-CAAR cohort at the University of Maryland for fostering the collaborative research environment and providing the facilities that made this work possible. Finally, we extend our gratitude to Red Cell Partners for their sponsorship of Aviva Prins.

\bibliographystyle{ACM-Reference-Format} 
\bibliography{references}

\end{document}